\def\year{2021}\relax
\documentclass[letterpaper]{article} 
\usepackage{aaai21}  
\usepackage{times}  
\usepackage{helvet} 
\usepackage{courier}  
\usepackage[hyphens]{url}  
\usepackage{graphicx} 
\usepackage{natbib}  
\usepackage{caption} 
\usepackage{amsmath}
\usepackage{amssymb}
\usepackage{bbm}
\usepackage{algorithm}
\usepackage{algpseudocode}
\usepackage{multirow}
\usepackage{makecell}
\usepackage{booktabs}

\newcommand{\etal}{\textit{et al}.}

\title{MeInGame: Create a Game Character Face from a Single Portrait}
\author{
    Jiangke Lin, \textsuperscript{\rm 1}
    Yi Yuan, \textsuperscript{\rm 1}\textsuperscript{\rm *}
    Zhengxia Zou \textsuperscript{\rm 2}\\
}
\affiliations{
    \textsuperscript{\rm 1} Netease Fuxi AI Lab \\
    \textsuperscript{\rm 2} University of Michigan \\
    \textsuperscript{\rm *} Corresponding Author \\
   linjiangke@corp.netease.com, yuanyi@corp.netease.com, zzhengxi@umich.edu
}

\begin{document}


\twocolumn[{
\renewcommand\twocolumn[1][]{#1}
\maketitle
\thispagestyle{empty}
\begin{center}
   \centering
   \includegraphics[width=\textwidth]{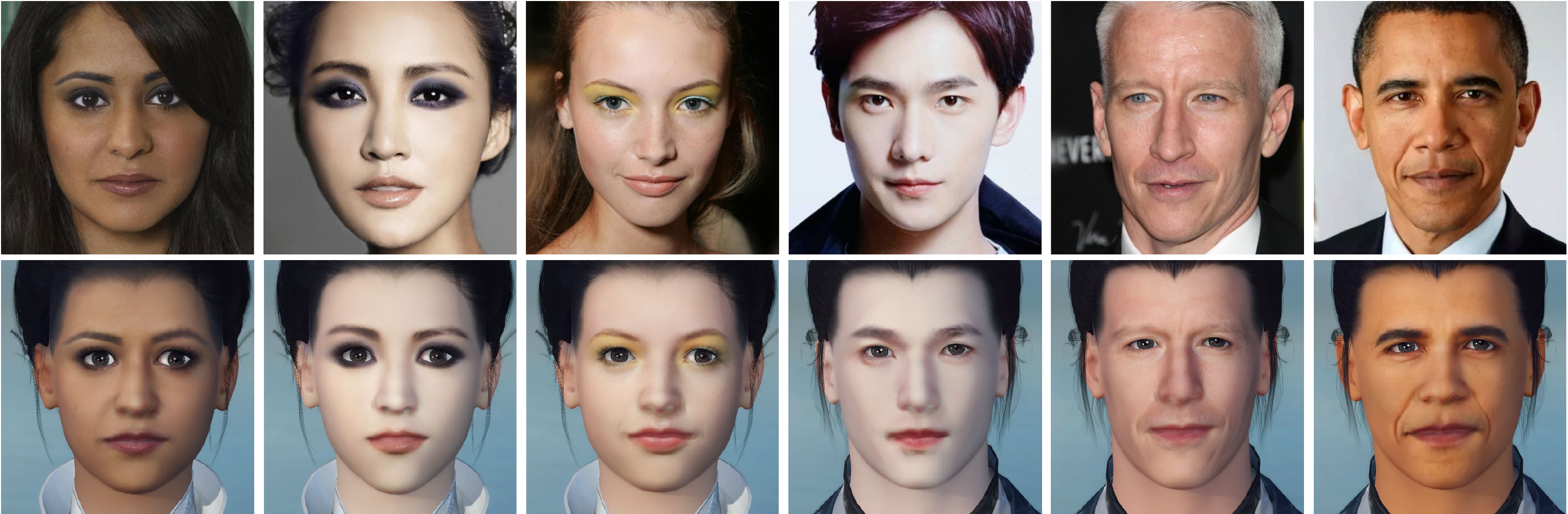}
   \captionof{figure}{First row: input portraits. Second row: in-game characters generated by our method. Our method is robust to lighting changes, shadows, and occlusions, and can faithfully restore personalized details like skin tone, makeup, and wrinkles.}
\end{center}
}]

\renewcommand{\thefootnote}{\fnsymbol{footnote}}
\footnotetext[2]{Code and dataset are available at \url{https://github.com/FuxiCV/MeInGame}.}
\renewcommand{\thefootnote}{\arabic{footnote}}

\begin{abstract}
Many deep learning based 3D face reconstruction methods have been proposed recently, however, few of them have applications in games. Current game character customization systems either require players to manually adjust considerable face attributes to obtain the desired face, or have limited freedom of facial shape and texture. In this paper, we propose an automatic character face creation method that predicts both facial shape and texture from a single portrait, and it can be integrated into most existing 3D games. Although 3D Morphable Face Model (3DMM) based methods can restore accurate 3D faces from single images, the topology of 3DMM mesh is different from the meshes used in most games. To acquire fidelity texture, existing methods require a large amount of face texture data for training, while building such datasets is time-consuming and laborious. Besides, such a dataset collected under laboratory conditions may not generalized well to in-the-wild situations. To tackle these problems, we propose 1) a low-cost facial texture acquisition method, 2) a shape transfer algorithm that can transform the shape of a 3DMM mesh to games, and 3) a new pipeline for training 3D game face reconstruction networks. The proposed method not only can produce detailed and vivid game characters similar to the input portrait, but can also eliminate the influence of lighting and occlusions. Experiments show that our method outperforms state-of-the-art methods used in games. 
\end{abstract}

\section{Introduction}

Due to the COVID-19 pandemic, people have to keep social distancing. Most conferences this year have been switched to online/virtual meetings. Recently, Dr. Joshua D. Eisenberg organized a special conference, Animal Crossing Artificial Intelligence Workshop (ACAI)\footnote{https://acaiworkshop.com/} in the Nintendo game \emph{Animal Crossing New Horizons}\footnote{https://www.animal-crossing.com/new-horizons/}, and has received great attention. Also, it is reported that this year's International Conference on Distributed Artificial Intelligence (DAI)\footnote{http://www.adai.ai/dai/2020/2020.html} will also be held in the game \emph{Justice}\footnote{https://n.163.com/} via cloud gaming techniques. As more and more social activities come to online instead of face-to-face, in this paper, we focus on game character auto-creation, which allows users to automatically create 3D avatars in the virtual game environment by simply upload a single portrait.

Many video games feature character creation systems, which allow players to create personalized characters. However, creating characters that look similar to the users themselves or their favorite celebrities is not an easy task and can be time-consuming even after considerable practices. For example, a player usually needs several hours of patience manually adjusting hundreds of parameters (e.g. face shape, eyes) to create a character similar to a specified portrait.

To improve a player's gaming experience, several approaches for game character auto-creation have emerged recently. Shi \etal proposed a character auto-creation method that allows users to upload single face images and automatically generate the corresponding face parameters~\cite{shi2019face}. However, such a method and its latest variant~\cite{shi2020fast} have limited freedom of the facial parameters and thus can not deal with buxom or slim faces very well. Besides, these methods do not take textures into account, which further limits their adaptability to different skin colors. Using a mobile device to scan a human face from multiple views to generate a 3D face model is another possible solution. The game NBA 2K\footnote{https://www.nba2k.com/} adopts this type of method. However, users have to wait several minutes before a character is created. Besides, this approach is not suitable for creating 3D faces for celebrities or anime characters, since their multi-view photos are hardly available for players.

To tackle the above problems, we propose a new method for automatic game character creation and a low-cost method for building a 3D face texture dataset. Given an input face photo, we first reconstruct a 3D face based on 3D Morphable Face Model (3DMM) and Convolutional Neural Networks (CNNs), then transfer the shape of the 3D face to the template game mesh. The proposed network takes in the face photo and the unwrapped coarse UV texture map as input, then predicts lighting coefficients and refined texture map. By utilizing the power of neural networks, the undesired lighting component and occlusions from the input can be effectively removed. As the rendering process of a typical game engine is not differentiable, we also take advantage of the differentiable rendering method to make gradients back-propagated from the rendering output to every module that requires parameter updating during training. In this way, all network components can be smoothly training in an end-to-end fashion. In addition to the differentiable rendering, we also design a new training pipeline based on semi-supervised learning in order to reduce the dependence of the training data. We use the paired data for supervised learning and the unlabeled data for self-supervised learning. Thus, our networks can be trained in a semi-supervised manner, reducing reliance on the pre-defined texture maps. Finally, by loading the generated face meshes and textures to the game environments, vivid in-game characters can be created for players. Various expressions can be further made on top of the created characters with blendshapes.

Our contributions are summarized as follows:

\begin{itemize}
  \item We propose a low-cost method for 3D face dataset creation. The dataset we created is balanced in race-and-gender, with both facial shape and texture created from in-the-wild images. We will make it publicly available after the paper got accepted.
  \item We propose a method to transfer the reconstructed 3DMM face shape to the game mesh which can be directly used in the game environment. The proposed method is independent of the mesh connectivity and is computationally efficient for practical usage.
  \item To eliminate the influence of lighting and occlusions, we train a neural network to predict an integral diffuse map from a single in-the-wild human face image under an adversarial training paradigm. 
\end{itemize}

\section{Related Work}

\subsection{3D Morphable Face Models}

Recovering 3D information from a single 2D image has long been a challenging but important task in computer vision. 3D Morphable Face Models (3DMM), as a group of representative methods for 3D face reconstruction, was originally proposed by Blanz and Vetter~\cite{blanz1999morphable}. In 3DMM and its recent variants~\cite{booth20163d,cao2013facewarehouse,gerig2018morphable,huber2016multiresolution,li2017learning}, the facial identity, expression and texture are approximated by low-dimensional representations from multiple face scans.

In a typical 3DMM model, given a set of facial identity coefficients $c_{i}$ and expression coefficients $c_{e}$, the face shape $S$ can be represented as follows:
\begin{equation}
  S = S_{mean} + c_{i}I_{base} + c_{e}E_{base},
\label{equation:3DMM}
\end{equation}
where $S_{mean}$ is the mean face shape, $I_{base}$ and $E_{base}$ are the PCA bases of identity and expression respectively.

\subsection{Facial Texture}

Face texture restoration aims to extract textures from input face photos. A popular solution for face texture restoration is to frame this process as an image-to-texture prediction based on supervised training. Zafeiriou \etal captured a large scale 3D face dataset~\cite{booth20163d}. They made the shape model public available but remain the texture information private. With such a private large scale dataset, Zafeiriou and his colleagues ~\cite{deng2018uv,zhou2019dense,gecer2019ganfit} produced good results on shape and texture restoration. 

However, many 3D face reconstruction methods do not involve the face texture restoration so far since it is expensive to capture face textures. On one hand, the data acquired in controlled environments can not be easily applied to in-the-wild situations. On the other hand, it is not easy to balance subjects from different races, which may lead to bias and lack of diversity in the dataset. In particular, most subjects in~\cite{booth20163d} are Caucasian, therefore, methods based on such a dataset may not be well generalized to Asians or Africans or other races. Recently, Yang \etal~\cite{yang2020facescape} spent six months collecting a face dataset named FaceScape from 938 people (mostly Chinese) and made this dataset publicly available. However, compare to~\cite{booth20163d}, FaceScape still has very limited scale.

\begin{figure*}[ht]
\centering
\includegraphics[width=\linewidth]{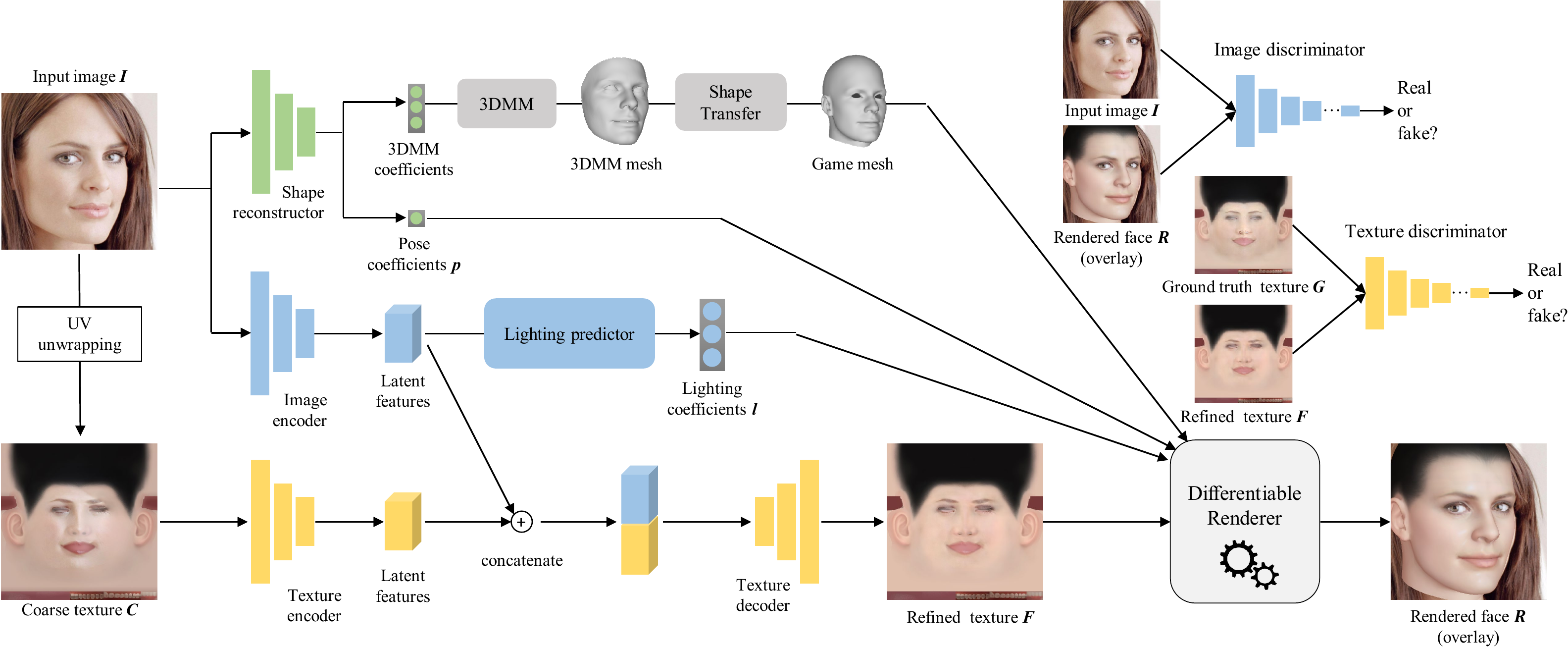}
\caption{An overview of our method. Given an input photo, a pre-trained shape reconstructor predicts the 3DMM and pose coefficients and a shape transfer module transforms the 3DMM shape to the game mesh while keeping their topology. Then, a coarse texture map is created by unwrapping the input image to UV space based on the game mesh. The texture is further refined by a set of encoder and decoder modules. We also introduce a lighting regressor to predict lighting coefficients from image features. Finally, the predicted shape, texture, together with the lighting coefficients, are fed to a differentiable renderer, and we force the rendered output similar to the input photo. Two discriminators are introduced to further improve the results.}
\label{fig:overall}
\end{figure*}

\subsection{Differentiable Rendering}

To address the dataset issue, differentiable rendering techniques~\cite{genova2018unsupervised,chen2019learning,liu2019soft} have been introduced to face reconstruction recently. With a differentiable renderer, an unsupervised/self-supervised loop can be designed where the predicted 3D objects (representing by meshes, point clouds, or voxels) can be effectively back-projected to 2D space, and thereby maximizes their similarity between the projection and the input image. However, even with a differentiable renderer, there is still a lack of constraints on the reconstructed 3D face information. Methods that used differentiable rendering~\cite{genova2018unsupervised,deng2019accurate} still rely on the prior knowledge of 3DMM, which are not fully unsupervised. Besides, such textures restored by 3DMM based methods cannot faithfully represent the personalized characteristics (e.g. makeup, moles) of the input portrait. Lin \etal~\cite{lin2020towards} recently propose to refine the textures from images by applying graph convolutional networks. While achieving high-fidelity results, these 3DMM based methods aim to reconstruct the 3D shape and texture for the face region rather than the whole head, which cannot be directly used for the game character creation. As a comparison, we aim to create the whole head model with a complete texture, whose shape and appearance are similar to the input.

\subsection{Shape Transfer}

Shape transfer aims to transfer shapes between two meshes. To generate a full head model instead of a front face only, we use shape transfer to transfer a 3DMM mesh to a head mesh with game topology. Non-rigid Iterative Closest Point (Non-rigid ICP) algorithm~\cite{amberg2007optimal} is the typical method for a shape transfer task, which performs iterative non-rigid registration between the surfaces of two meshes. Usually, non-rigid ICP and its variants has good performance on meshes with regular topology. However, such methods normally take several seconds to complete a transfer, which is not fast enough for our task.


\section{Approach}

Fig.~\ref{fig:overall} shows an overview of the proposed method. We frame the reconstruction of the face shape and texture as a self-supervised facial similarity measurement problem. With the help of differentiable rendering, we design a rendering loop and force the 2D face rendering from the predicted shape and texture similar to the input face photo.

Our method consists of several trainable sub-networks. The image encoder takes in the face image as input and generates latent features. The image features are then flattened and fed to the lighting regressor - a lightweight network consists of several fully-connected layers and predicts lighting coefficients (light direction, ambient, diffuse and specular color).  Similar to image encoder, we introduce a texture encoder. The features of the input image and coarse texture map are concatenated together and then fed into the texture decoder, producing the refined texture map. With the game mesh, the refined texture map, pose, and lighting coefficients, we use the differentiable renderer~\cite{ravi2020pytorch3d} to render the face mesh to a 2D image and enforce this image to be similar with the input face photo. To further improve the results, we also introduce two discriminators, one for the rendered face image and another for the generated face texture maps.

\subsection{Dataset Creation}\label{sec:data_create}

Here we introduce our low-cost 3D face dataset creation method. Unlike other methods that require multi-view images of subjects, which are difficult to capture, our method only uses single view images and is easily acquired. With such a method, we, therefore, create a Race-and-Gender-Balance (RGB) dataset and name it "RGB 3D face dataset".

The dataset creation includes the following steps:

\begin{itemize}
  \item [\romannumeral1.] Given an input face image, detect the skin region by using a pre-trained face segmentation network.
  \item [\romannumeral2.] Compute the mean color of the input face skin and transfer the mean skin color to the template texture map (provided by the game developer).
  \item [\romannumeral3.] Unwrapping the input face image to UV space according to the deformed game head mesh.
  \item [\romannumeral4.] Blend the unwrapped image with the template texture map using Poisson blending. Remove the non-skin regions such as hair and glasses, and use symmetry to patch up the occluded regions when possible.
\end{itemize}

\begin{figure}[h]
\centering
\includegraphics[width=\columnwidth]{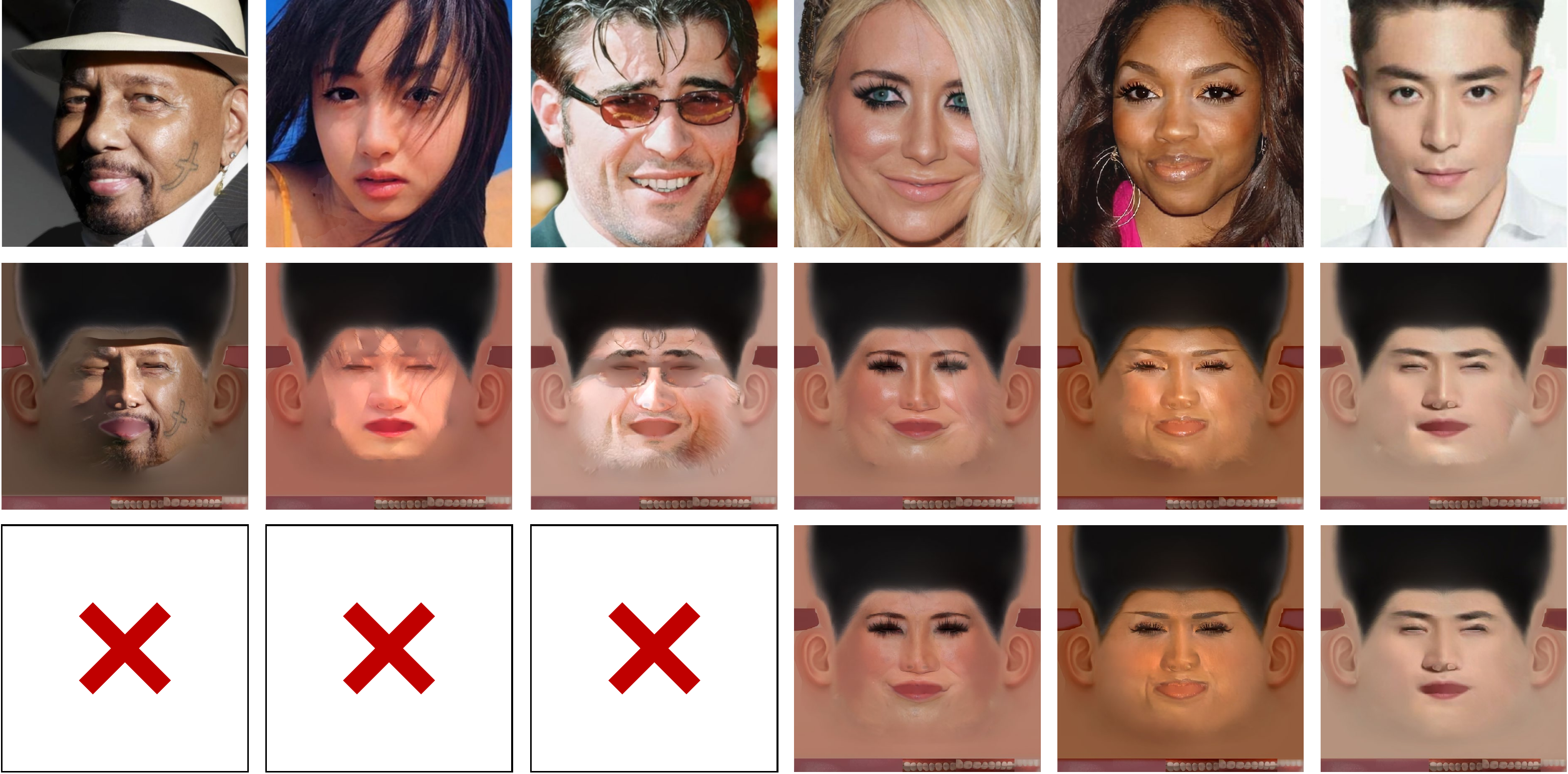}
\caption{some examples of generated texture maps. the first row: input images; second row: coarse texture maps. we select those good quality texture maps (the three on the right) to further create ground truth, as shown in the last row.}
\label{fig:uv_creation}
\end{figure}

Fig.~\ref{fig:uv_creation} shows some texture maps created by using the above method. The texture maps with good quality are chosen for further refinement. We manually edit the coarse texture map by using an image editing tool (e.g. \emph{PhotoShop}) to fix the shadows and highlights. Since we can control the quality of the generated texture, the workload of manual repair is very small, and each face only takes a few minutes to complete the refinement.

\subsection{Face Shape Reconstruction}\label{sec:shape_recon}

The very first step of our method is to predict the 3DMM shape and pose coefficients from an input image. In this paper, We adopt the 3DMM coefficient regressor in~\cite{deng2019accurate} for face shape reconstruction, but other 3D face reconstruction methods will also work. Given a 2D image, we aim to predict a 257-dimensional vector $(c_{i}, c_{e}, c_{t}, p, l)\in\mathbb{R}^{257}$, where $c_{i}\in\mathbb{R}^{80}$, $c_{e}\in\mathbb{R}^{64}$ and $c_{t}\in\mathbb{R}^{80}$ represent the 3DMM identity, expression and texture coefficients respectively. $p\in\mathbb{R}^{6}$ is the face pose and $l\in\mathbb{R}^{27}$ represents the lightings. With the predicted coefficients, the face vertices' 3D positions $S$ can be computed based on Eq.~\ref{equation:3DMM}.

\subsection{Shape Transfer}\label{sec:shapetransfer}

The shape transfer module aims to transfer the reconstructed 3DMM mesh to the game mesh. We design our shape transfer module based on Radial Basis Function (RBF) interpolation~\cite{de2007mesh}.

Specifically, RBF defines a series of basis functions as follows:
\begin{equation}
\Phi_i (x) = \varphi_i (\|x - x^\prime\|),
\end{equation}
where $x^\prime$ represents the center of $\varphi$ and $\|x - x^\prime\|$ denotes the Euclidean distance between the input point $x$ and center $x^\prime$.

In this way, given a point $x$ and a set of RBF basis functions, the value of $f(x)$ can be computed using RBF interpolation:
\begin{equation}
  f(x)=\sum_{i=0}^{n}{w_i \varphi(\|x - x^\prime_i\|)}
\end{equation}
where $w_i$ represents the weight of each basis, which are the unknowns to be solved.

In the scenario of mesh shape transfer, we first manually specified 68 landmark pairs between the 3DMM template mesh and the game template mash, similar to those in dlib~\cite{dlib09}. In the following, we donote those landmarks on template mesh as original face landmarks.

Then we set the centers of the basis functions $x^\prime$ as the original face landmarks $\widetilde{L_g}$ on the game mesh. The input $x$ is set to the original position of a game mesh vertex and the value $f(x)$ we get is the offset of the transfer. In this way, the vertex's new position can be computed as $x+f(x)$. To determine the weights $w_i$, we propose to solve a linear least-squares problem by minimizing the distance of the face landmark pairs between the game mesh and 3DMM mesh. For more details about the shape transfer, please refer to our supplementary materials and code.

\subsection{Loss Functions}

We design our loss functions to minimize the distance between the rendered face image and the input face photo, and the distance between the refined texture map and the ground truth texture map. Within the rendering loop, we design four types of loss functions, i.e. the pixel loss, the perceptual loss, the skin regularization loss, and the adversarial loss, to measure the facial similarity from both global appearance and local details.

\subsubsection{Pixel Loss}

We compute our pixel loss on both of the rendered image space and the texture UV space.

For the rendered image $R$, the loss is computed between $R$ and its corresponding input image $I$. We define the loss as the pixel-wise L1 distance between the two images:
\begin{equation}
  \mathcal{L}_{rec}(I, R)=\frac{\sum_{i \in \mathcal{M}}{\|I_i - R_i\|_1}}{|\mathcal{M}_{2d}|}
\end{equation}
where $i$ is the pixel index, $\mathcal{M}_{2d}$ is the skin region mask obtained by the face segmentation network in 2D image space.

For the pixel loss on UV space, we define this loss as the L1 distance between the refined texture map $F$ and the ground truth texture map $G$:
\begin{eqnarray}
  \mathcal{L}_{tex}(F, G)= \frac{1}{N}\sum_{i=0}^{N}{\|F_i - G_i\|_1}
\end{eqnarray}
where $i$ is the pixel index and $N$ is the number of pixels.

\subsubsection{Perceptual Loss}

We following Nazeri \etal~\cite{nazeri2019edgeconnect} and design two losses in perception level, i.e. perceptual loss $\mathcal{L}_{perc}$ and style loss $\mathcal{L}_{style}$. Given a pair of images $x$ and $x^\prime$, the perceptual loss is defined by the distance between their activation maps of a pre-trained network (e.g., VGG-19):
\begin{equation}
  \mathcal{L}_{perc}(x,x^\prime)=\mathbb{E}\left[\sum_{i}\frac{1}{N_i}\|\phi_i (x) - \phi_i (x^\prime) \|_1\right]
\end{equation}
where $\phi_i$ is the activation map of the $i^{th}$ layer of the network. The style loss is on the other hand defined on the covariance matrices of the activation maps:
\begin{equation}
  \mathcal{L}_{sty}(x,x^\prime)=\mathbb{E}_j \left[\|G_j^\phi(x) - G_j^\phi(x^\prime)\|_1\right]
\end{equation}
where $G_j^\phi(x)$ is a Gram matrix constructed from activation maps $\phi_j$.

We compute the above two losses above on both the face images and texture maps.

\subsubsection{Skin Regularization Loss}

To produce a constant skin tone across the whole face and remove highlights and shadows, we conduct two losses to regularize the face skin, namely a ``symmetric loss'' and a ``standard-deviation loss''. Unlike previous works~\cite{tewari2018self,deng2019accurate} that apply skin regularization directly on vertex color, we impose the penalties on the Gaussian blurred texture map. This is based on a fact that some personalized details (e.g. a mole) are not always symmetric and not related to skin tone.

We define the symmetric loss as follows:
\begin{equation}
  \mathcal{L}_{sym}(F')=\frac{2}{N_U \times N_V}\sum_{i}^{N_V}{\sum_{j}^{N_U/2}{F'_{i,j} - F'_{i, N_U-j}}}
\end{equation}
where $F'$ is the Gaussian blurred refined texture map $F$. $N_U$ and $N_V$ are the numbers of columns and rows of the texture map respectively.

We define the skin standard-deviation loss as follows:
\begin{equation}
  \mathcal{L}_{std}(F')=\sqrt{\frac{1}{|\mathcal{M}_{uv}|}\sum_{i \in \mathcal{M}_{uv}}(F'_i - \bar{F'})^2}
\end{equation}
where the $\bar{F'}$ is the mean value of $F'$, $\mathcal{M}_{uv}$ is the skin region mask in UV space and $i$ is the pixel index.

\subsubsection{Adversarial Loss}

To further improve the fidelity of the reconstruction, we also use adversarial losses during the training. We introduce two discriminators, one for the rendered face and one for the generated UV texture maps, respectively. We train the discriminators to tell whether the generated outputs are real or fake, at the same time, we train other parts of our networks to fool the discriminators. The objective functions of the adversarial training are defined are follows:
\begin{equation}
\begin{split}
  \mathcal{L}_{gen}&=\mathbb{E}[\log D_i(x^\prime)]\\  
    \mathcal{L}_{dis}&=\mathbb{E}[\log D_i(x)]+\mathbb{E}[\log (1-D_i(x^\prime))],
\end{split}
\end{equation}
where $D_i\in\{D_{img}, D_{tex}\}$ are the discriminators for image and texture map separately.

\subsubsection{Final Loss Function}

By combining all the above defined losses, our final loss functions can be written as follows:
\begin{equation}
\begin{aligned}
  \mathcal{L}_G=&\lambda_{l1}(\mathcal{L}_{ren}(I, R) + \mathcal{L}_{tex}(F, G))\\
              &+\lambda_{perc}(\mathcal{L}_{perc}(I, R) + \mathcal{L}_{perc}(F, G))\\
              &+\lambda_{sty}(\mathcal{L}_{sty}(I, R) + \mathcal{L}_{sty}(F, G))\\
              &+\lambda_{sym}\mathcal{L}_{sym}(F') + \lambda_{std}\mathcal{L}_{std}(F')\\
              &+\lambda_{adv}(\mathcal{L}_{gen}(R|D_{img}) + \mathcal{L}_{gen}(F|D_{tex})),
\end{aligned}
\end{equation}
\begin{equation}
  \mathcal{L}_D=\lambda_{adv}(\mathcal{L}_{dis}(I,R|D_{img})+\mathcal{L}_{dis}(F,G|D_{tex})),
\end{equation}
where $\mathcal{L}_G$ is the loss for training the image encoder, texture map encoder, light regressor and texture map decoder. $\mathcal{L}_D$ is the loss for training the discriminators. $\lambda$s are the corresponding weights to balance the different loss terms.

During the training, we aim to solve the following mini-max optimization problem: $\min_G\max_D \mathcal{L}_G + \mathcal{L}_D$. In this way, all the network components to be optimized can be trained in an end-to-end fashion. For these texture maps that do not have paired ground truth data, we simply ignore corresponding loss items during the training process.

\begin{figure*}[t]
  \centering
  \includegraphics[width=0.8\linewidth]{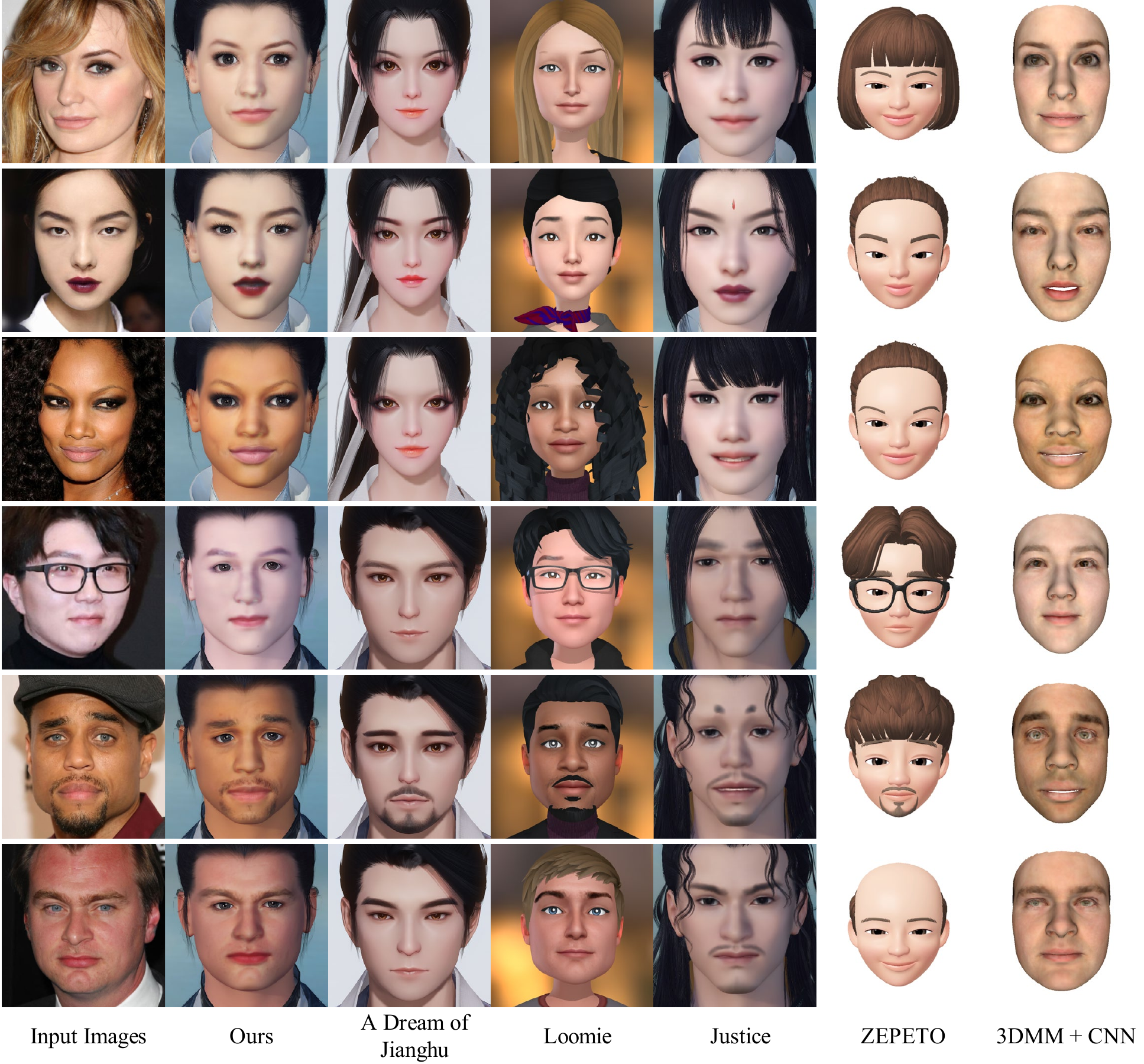}
   \caption{Comparison with other methods used in games: \emph{A Dream of Jianghu}, \emph{Loomie}, \emph{Justice}~\cite{shi2020fast}, \emph{ZEPETO}. We also show the results of a 3DMM-based method~\cite{deng2019accurate} in the last column as references.}
\label{fig:compare}
\end{figure*}

\section{Implementation Details}

We use the Basel Face Model~\cite{paysan20093d} as our 3DMM. We adopt the pre-trained network from~\cite{deng2019accurate} to predict the 3DMM and pose coefficients. The 3DMM face we used contains 35,709 vertices and 70,789 faces. We employ the head mesh from a real game, which contains 8,520 vertices and 16,020 faces. The facial segmentation network we used is from~\cite{shi2019face}.

We use the CelebA-HQ dataset~\cite{karras2017progressive} to create our dataset. During the UV unwrapping process, we set the resolution of the face images and the unwrapped texture maps to $512 \times 512$ and $1,024 \times 1,024$ respectively. The dataset we created consists of six subsets: \{Caucasian, Asian and African\} $\times$ \{female and male\}. Each subset contains 400 texture maps. From each subset, we randomly select 300 for training, 50 for evaluation, and 50 for testing.

Note that the game mesh we produced only corresponds to an enclosed head. Hair, beard, eyeballs, etc. are not considered in our method. This is because, in many games, heads, hair, etc. are topologically independent modules. Players can freely change various hairstyles (including long hair, etc.) and other decorations without affecting the topological structure of the head.

We use grid-search on the loss weights from 0.001 to 10, and we select the best configuration based on the overall loss on the validation set as well as quantitative comparison. The weights of loss terms are finally set as follows: $\lambda_{l1}=3, \lambda_{perc}=1$, $\lambda_{sty}=1$, $\lambda_{sym}=0.1$, $\lambda_{std}=3$, $\lambda_{adv}=0.001$. The learning rate is set to 0.0001, we use the Adam optimizer and train our networks for 50 epochs.

We run our experiments on an Intel i7 CPU and an NVIDIA 1080Ti GPU, with PyTorch3D (v0.2.0) and its dependencies. Given a portrait and coarse texture map, our network only takes 0.4s to produce a $1,024 \times 1,024$ refined texture map.


\section{Experimental Results}

\subsection{Qualitative Comparison}

We compare our method with some other state-of-the-art game character auto-creation methods/systems, including the character customization systems in \emph{A Dream of Jianghu}\footnote{https://jianghu.163.com}, \emph{Loomie}\footnote{https://loomai.com}, \emph{Justice}\footnote{https://n.163.com} (which is based on the method of~\cite{shi2020fast}), and \emph{ZEPETO}\footnote{https://zepeto.me}.

As shown in Fig.~\ref{fig:compare}, our results are more similar to the input images than the other results in both of the face shape and appearance. The faces reconstructed by \emph{Justice}~\cite{shi2020fast}, \emph{A Dream of Jianghu}, and \emph{ZEPETO} have limited shape variance, and also fail to recover the textures of the input images. \emph{Loomie} restores both facial shape and texture, but it cannot handle difficult lighting conditions (e.g. highlights), occluded regions (e.g. eyebrow of the first example), and personalized details (e.g. makeup). 

We also compare with the state-of-the-art 3DMM based method~\cite{deng2019accurate} in Fig.~\ref{fig:compare} which applies CNNs to reconstruct 3DMM faces from single images. We can see that the 3DMM only models facial features and does not include a complete head model as well as textures, making it difficult to be directly used in the game environments.

For more comparisons please refer to our supplementary materials.

\subsection{Quantitative Comparison}

Since it is hard to acquire ground truth data of game characters, we perform a user study between our results and others. We invited 30 people to conduct the evaluation. Each person was assigned with 480 groups of results. Each group of results included a portrait, our result, and a result from others. Participants were asked to choose a better one from the two results by comparing them with the reference portrait. We believe the user reported score reflects the quality of the results more faithfully than other indirect metrics. The statistics of the user study results are shown in Tab.~\ref{tab:userstudy}. The user study shows that ours are significantly better than others.

\begin{table}[t]
\centering
\begin{tabular}{c|cccc}
\toprule
      & \makecell[c]{A Dream\\of Jianghu} & Loomie & Justice & ZEPETO\\
\midrule
\makecell[c]{Others\\/\\Ours}   & \makecell[c]{0.0075\\/\\0.9925} & \makecell[c]{0.0394\\/\\0.9606} & \makecell[c]{0.0336\\/\\0.9644} & \makecell[c]{0.0081\\/\\0.9919} \\
\bottomrule
\end{tabular}
\caption{User preferences of the character auto-creation methods featured in different games.}
\label{tab:userstudy}
\end{table}

In addition to the user study on the final reconstructions, we also compare our method with Deng \etal~\cite{deng2018uv} on the quality of restored textures. Deng \etal did not take lighting or occlusions into consideration, which makes their method more like image inpainting than texture reconstruction. We compute Peak Signal-to-Noise Ratio (PSNR) and Structural Similarity Index Measure (SSIM) metric between the refined texture maps and the ground truth texture maps. The scores are shown in Tab.~\ref{tab:compare}. Note that Deng \etal~\cite{deng2018uv} reported their results on WildUV, a dataset similar to ours which is also constructed from an in-the-wild dataset (UMD video dataset~\cite{bansal2017s}). A direct comparison with our results on RGB 3D Face could be unfair to some extent. Nevertheless, here we still list their result as a reference. 

\begin{table}[t]
\centering
\begin{tabular}{c|cc}
\toprule
                         & PSNR & SSIM \\
\midrule
WildUV~\cite{deng2018uv} & 22.9 & 0.887 \\
\midrule
RGB 3D Face (ours)       & 24.2 & 0.905 \\
\bottomrule
\end{tabular}
\caption{PSNR and SSIM of our method and Deng's method on RGB 3D dataset and WildUV dataset.}
\label{tab:compare}
\end{table}




\subsection{Ablation Study}

\begin{figure}[h]
  \centering
  \includegraphics[width=\linewidth]{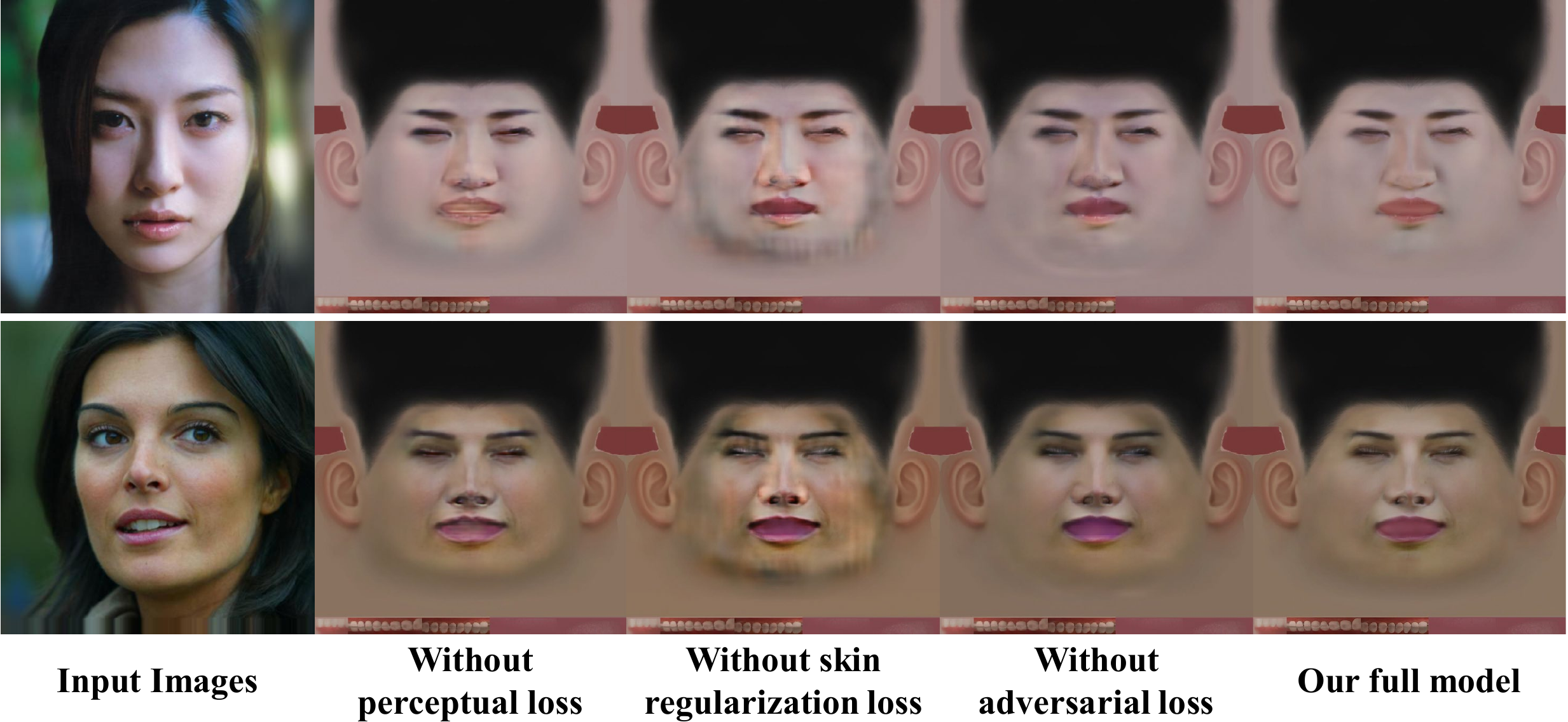}
  \caption{Ablation study on different loss functions.}
\label{fig:ablation}
\end{figure}

To evaluate the contribution of different loss functions, we conduct ablation studies on perceptual loss, skin regularization loss, and adversarial loss. Some examples are demonstrated in Fig.~\ref{fig:ablation}, the full model generates more realistic texture maps.

We also compute the PSNR and SSIM metrics, the scores are shown in Tab.~\ref{tab:ablation}.

\begin{table}[h]
\centering\setlength{\tabcolsep}{5pt}{
\begin{tabular}{c|c|cccc}
\toprule
\multirow{4}*{\rotatebox{90}{Losses}}
& Pixel               & \checkmark & \checkmark & \checkmark & \checkmark \\
& Perceptual          &            & \checkmark & \checkmark & \checkmark \\
& Skin Regularization & \checkmark &            & \checkmark & \checkmark \\
& Adversarial         & \checkmark & \checkmark &            & \checkmark \\
\midrule
\multicolumn{2}{c|}{PSNR $\uparrow$} & 24.10 & 23.83 & 24.13 & \textbf{24.21} \\
\multicolumn{2}{c|}{SSIM $\uparrow$} & 0.901 & 0.898 & 0.904 & \textbf{0.905} \\
\bottomrule
\end{tabular}}
\caption{Metrics on ablation study.}
\label{tab:ablation}
\end{table}

\subsection{Discussion}

\begin{figure}[h]
  \centering
  \includegraphics[width=0.85\linewidth]{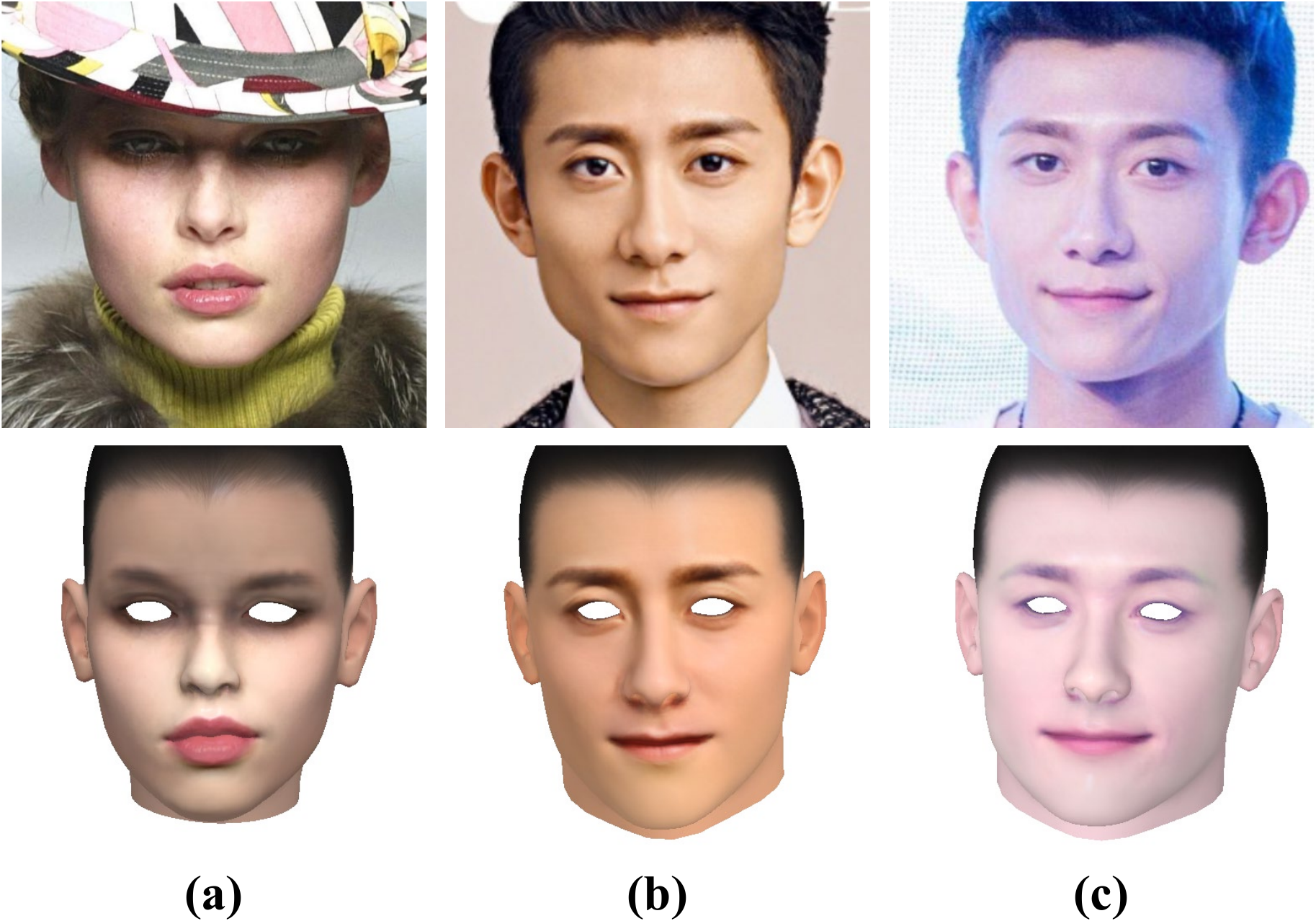}
  \caption{Special cases of the proposed method, a) shadows caused by hats that cannot be modeled by the differentiable renderer, b) and c) show the same person and the result under severe lighting changes.}
\label{fig:limitation}
\end{figure}

Although we achieve higher accuracy than other methods in quantitative and qualitative metrics, our method still has some limitations. As shown in Fig.~\ref{fig:limitation} (a), when there are heavy occlusions (e.g., the hat), our method fails to produce faithful results since the renderer fails to model the shadow created by the objects outside the head mesh.

Fig.~\ref{fig:limitation} (b, c) show the results from two portraits of the same person under severe lighting changes. Given (b) or (c) alone, either of the results looks good. Theoretically, it should produce similar results for the same person. However, the results are affected by lights of different colors.

\section{Conclusion}

In this paper, we present a novel method for automatic creation of game character faces. Our method produces character faces similar to the input photo in terms of both face shape and textures. Considering it is expensive to build 3D face datasets with both shape and texture, we propose a low-cost alternative to generate the data we need for training. We introduce a neural network that takes in a face image and a coarse texture map as inputs and predicts a refined texture map as well as lighting coefficients. The highlights, shadows, and occlusions are removed from the refined texture map and personalized details are preserved. We evaluate our method quantitatively and qualitatively. Experiments demonstrate our method outperforms the existing methods applied in games.

\bibliography{uv_inpainting.bib}
\end{document}